\documentclass[a4paper, 10pt, conference]{ieeeconf}      

\IEEEoverridecommandlockouts                              

\title{\LARGE \bf
	DeepCLR: Correspondence-Less Architecture for\\Deep End-to-End Point Cloud Registration 
}

\author{Markus Horn, Nico Engel, Vasileios Belagiannis, Michael Buchholz and Klaus Dietmayer
	\thanks{All authors are with the Institute of Measurement, Control and Microtechnology, Ulm University,
		Albert-Einstein-Allee 41, 89081 Ulm, Germany
		{\tt\small \{firstname\}.\{lastname\}@uni-ulm.de}}%
}

\usepackage[
backend=biber,
bibencoding=utf8,
style=ieee,
firstinits=true,
doi=false,
isbn=false,
url=false,
sorting=none
]{biblatex}
\usepackage{amsfonts}
\usepackage{amsmath}
\usepackage{bm}
\usepackage{booktabs}
\usepackage{graphicx}
\usepackage{hyperref}
\usepackage{multirow}
\usepackage{siunitx}
\usepackage{subfigure}
\usepackage{tikz}
\usepackage{tikzscale}
\usepackage{threeparttablex}

\graphicspath{{img/}}
\DeclareMathOperator*{\MAX}{MAX}

\newcommand\copyrighttext{\footnotesize \textcopyright~2020 IEEE. Personal use of this material is permitted. Permission from IEEE must be obtained for all other uses, in any current or future media, including reprinting/republishing this material for advertising or promotional purposes, creating new collective works, for resale or redistribution to servers or lists, or reuse of any copyrighted component of this work in other works.
DOI: \href{https://doi.org/10.1109/ITSC45102.2020.9294279}{10.1109/ITSC45102.2020.9294279}
}%

\newcommand\copyrightnotice{%
	\begin{tikzpicture}[remember picture,overlay]
	\node[anchor=south,xshift=9pt,yshift=10pt] at (current page.south) {\fbox{\parbox{\dimexpr\textwidth-\fboxsep-\fboxrule\relax}{\copyrighttext}}};
	\end{tikzpicture}%
}

\begin{document}
	
	\maketitle
	\thispagestyle{empty}
	\pagestyle{empty}

	\begin{abstract}
		This work addresses the problem of point cloud registration using deep neural networks.
		We propose an approach to predict the alignment between two point clouds with overlapping data content, but displaced origins.
		Such point clouds originate, for example, from consecutive measurements of a LiDAR mounted on a moving platform.
		The main difficulty in deep registration of raw point clouds is the fusion of template and source point cloud.
		Our proposed architecture applies flow embedding to tackle this problem, which generates features that describe the motion of each template point.
		These features are then used to predict the alignment in an end-to-end fashion without extracting explicit point correspondences between both input clouds.
		We rely on the KITTI odometry and ModelNet40 datasets for evaluating our method on various point distributions.
		Our approach achieves state-of-the-art accuracy and the lowest run-time of the compared methods.
	\end{abstract}
	
	
	\section{INTRODUCTION}
	\label{sec:introduction}
	
\begin{figure}[t]
		\centering
		\includegraphics[width=0.95\linewidth, trim={0 14cm 5cm 0}, clip]{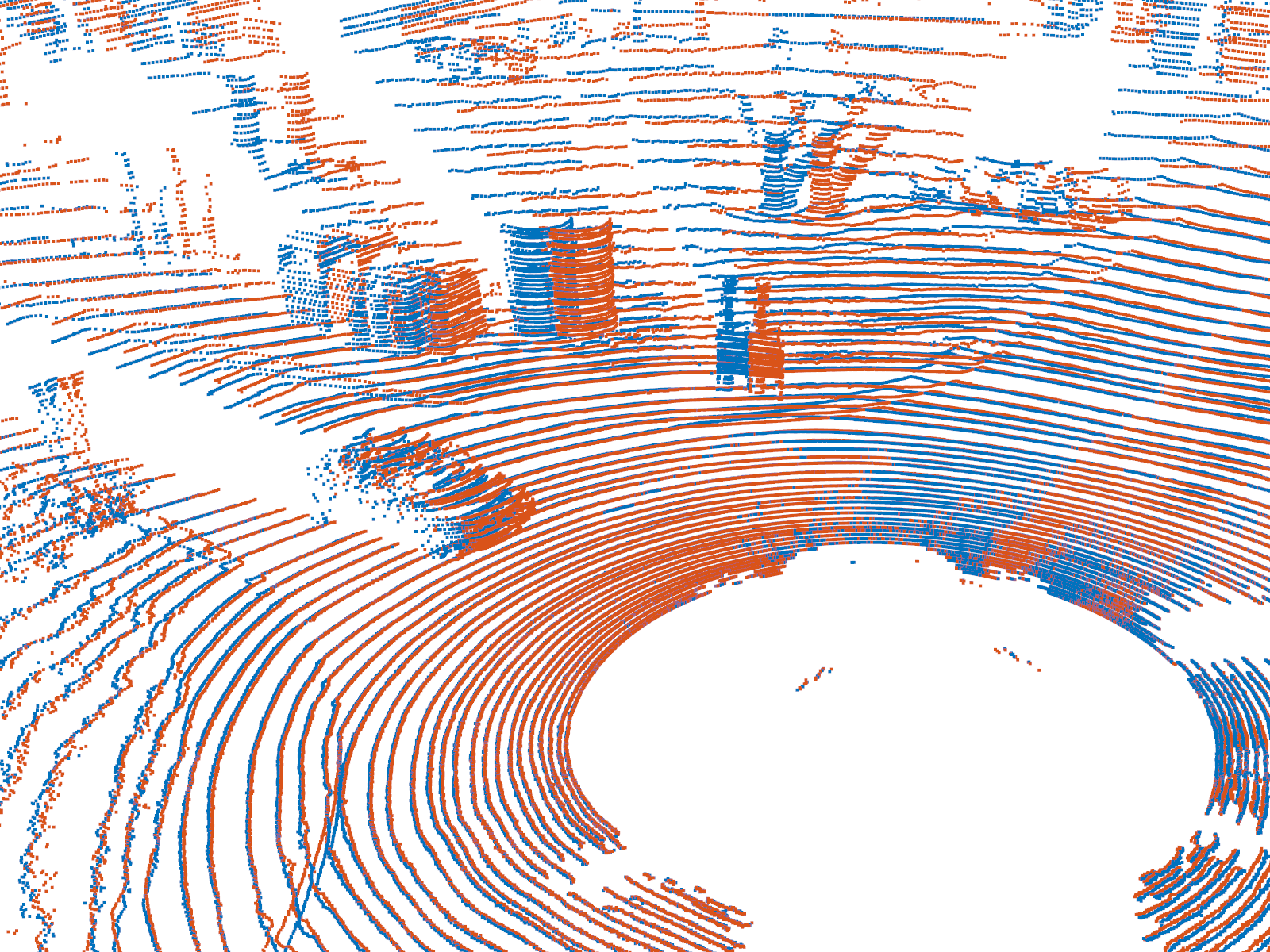}
		\caption{KITTI odometry example \cite{geiger2012autonomous}. LiDAR point cloud measurements of two consecutive frames $t$ in blue and $t+1$ in red. The vehicle is driving a left turn, causing a high rotation between the measurements.}
		\label{fig:kitti_crossing}
	\end{figure}

	\copyrightnotice
	The objective of point cloud registration (PCR) is the alignment of two point clouds, called template and source, by estimating their relative transformation.
	Point clouds are obtained from various sources such as LiDAR (Light Detection and Ranging) sensors or \mbox{RGB-D} cameras.
	In case a LiDAR is mounted on a robot or vehicle, we can apply PCR to obtain the sensor motion by estimating the relative transformation between measurements of two consecutive timesteps.
	This application of PCR called LiDAR odometry (LO) is one of the main tasks for our proposed architecture.
	An example from the KITTI odometry dataset \cite{geiger2012autonomous} is shown in Fig.~\ref{fig:kitti_crossing}.
	
	The most common algorithm for point cloud registration is Iterative Closest Point (ICP) \cite{besl1992method}.
	Since its publication in 1992, many variants with modified distance metrics, filtering steps and association methods have been introduced \cite{pomerleau2013comparing}.
	In the original work, Basler et al. prove that ICP achieves local convergence, which makes it suitable for fine registration.
	For larger translations and rotations, registration algorithms like Fast Point Feature Histograms (FPFH) \cite{rusu2009fast} as coarse pre-registration are crucial for reaching the global transformation error minimum.
	While ICP is a point based algorithm that uses the coordinates of input points for registration, FPFH is a feature based approach that subsamples the point clouds and performs registration based on local features.
	Nevertheless, even for small transformations both methods still struggle on noisy data and large numbers of points.
	However, continuing the example of LiDAR odometry from above, this problem requires an algorithm that can handle large point clouds with high noise.
	
	Especially for feature based approaches, it seems promising to apply deep learning for feature extraction, since neural networks can model higher order dependencies \cite{yew20183dfeatnet}.
	Recent publications for deep learning based PCR \cite{wang2019deeppco, li2019lonet} apply projections to generate pixel representations.
	This makes it possible to apply methods based on convolutional neural networks (CNN) like PoseNet \cite{kendall2015posenet}, originally introduced for camera images.
	While it is possible to project LiDAR point clouds to 2D panoramic view depth images \cite{wang2019deeppco} from the sensor origin, such a projection is not possible for arbitrary point clouds without loosing information.
	ModelNet40 \cite{wu20153d} is an example for such data where no projection is possible, since any projection would cause self-occlusions and therefore missing information.
	Thus, our goal is to develop an algorithm that processes raw point clouds without applying projections for ensuring complete data utilization.
	
	Recent advances in deep learning like PointNet \cite{qi2017pointnet} enable direct processing of unsorted point clouds without prior projection.
	Many architectures for deep registration of raw point clouds are based on this concept.
	DeepLocalization \cite{engel2019landmark} and PointNetLK \cite{aoki2019pointnetlk} use global PointNet feature vectors extracted directly for registration.
	However, these methods face problems with large point clouds like LiDAR measurements, since the global feature vectors do not hold sufficient information for a precise registration.
	An alternative approach is to use PointNet for extracting keypoints and correspondence weights, as in 3DFeat-Net \cite{yew20183dfeatnet} or DeepVCP \cite{lu2019deepvcp}.
	Since these methods enforce explicit point correspondences and thereby constrain the decision space, they do not exploit the full potential of deep learning.
	Furthermore, they rely on a subsequent algorithm for estimating the registration which leads to higher run-times.
	
	Our proposed architecture "DeepCLR" (Correspondence-Less Registration) addresses these problems by creating feature vectors describing the local point neighborhood as an alternative to global feature vectors for encoding the input.
	Instead of extracting explicit correspondences, we fuse the point clouds and create flow features for each template point cloud sample, which describe the estimated motion of each point between template and source.
	Finally, the resulting point cloud with added flow features is used to predict the alignment in an end-to-end fashion.
	
	To summarize, the main contribution of this work is an architecture which 1) directly processes unsorted point clouds of arbitrary size without applying a projection, 2) estimates the transformation in an end-to-end fashion without extracting explicit correspondences and 3) is able to perform the registration sufficiently fast for online processing. We evaluate our architecture using the KITTI odometry \cite{geiger2012autonomous} and ModelNet40 \cite{wu20153d} datasets.

	\section{RELATED WORK}
	\label{sec:related_work}
	
	Iterative Closest Point (ICP) \cite{besl1992method} and its variants \cite{pomerleau2013comparing} all involve data association for extracting the closest source point to each template point, and error minimization for iterative transformation refinement.
	The error calculation for associated point pairs can be based on the point-to-point or point-to-plane distance.
	Generized ICP (G-ICP) \cite{segal2010generalized} uses a probabilistic framework based on a plane-to-plane distance metric.
	Since ICP only guarantees local convergence, other registration methods are necessary as pre-registration for large transformations.
	The feature based FPFH \cite{rusu2009fast} calculates feature histograms for each point from its local neighborhood, which are then used for extracting corresponding point pairs.
	
	A wide range of deep learning based algorithms for point cloud registration focus on LiDAR point clouds.
	Due to their measuring principle, it is obvious to apply cylindrical or spherical projections for generating 2D panoramic-view depth images.
	This advantage is used by DeepPCO \cite{wang2019deeppco} and LO-Net \cite{li2019lonet} for LiDAR odometry estimation using CNNs.
	DeepPCO uses a dual-branch architecture in order to predict translation and rotation separately.
	In LO-Net, the depth image representation is used to estimate the surface normals for each pixel, which are then provided as additional input.
	Since only measurements of static objects can be used for odometry estimation, LO-Net predicts a mask to suppress measurements of dynamic objects.
	Nevertheless, these approaches for LiDAR odometry cannot be applied directly on registration of arbitrary point clouds, as explained in Section~\ref{sec:introduction}.
	Since our architecture is based on PointNet++ \cite{qi2017pointnetpp} we are able to process raw point clouds from any distribution without prior projection or rasterization.
	Thus, we are not limited on LiDAR measurements.
	
	For direct processing of point clouds without projections, PointNet \cite{qi2017pointnet} or PointNet++ \cite{qi2017pointnetpp} provide powerful foundations.
	PointNet is used in DeepLocalization \cite{engel2019landmark} and PointNetLK \cite{aoki2019pointnetlk} for generating feature vectors from template and source point clouds.
	These features are concatenated in DeepLocalization and passed to fully connected layers for pose prediction.
	In PointNetLK, instead of concatenating feature vectors, these are used to calculate gradients for iterative transformation refinement.
	Both methods are developed for small point clouds.
	Unfortunately, PointNet fails to generate descriptive feature vectors for accurate registration of large point clouds like LiDAR measurements.
	
	Another point based approach using PointNet++ is presented by DeepVCP \cite{lu2019deepvcp}.
	Lu et al. subsample and group input points for calculating features, which describe the local surrounding structure of each sample.
	For merging both subsampled point clouds in the feature embedding, they collect all surrounding source points for each template point, concatenate the features and feed them into mini-PointNets similar to PointNet++.
	These fused features are used to predict correspondence weights for multiple point pairs, which are then processed afterwards to estimate the relative transformation between template and source point cloud.
	This feature embedding approach is similar to the flow embedding of FlowNet3D \cite{liu2019flownet3d}, which is used for point cloud scene flow estimation.
	
	The concept of our proposed architecture is based on a similar method for fusing the point clouds.
	Our major innovation is the end-to-end approach, which means we do not predict any explicit correspondences or correspondence weights between template and source point clouds.
	Instead of using the embedding result for predicting correspondence weights, we use the PointNet architecture to infer the transformation directly from the embedded feature representation.
	Thus, the network can directly predict the registration without being forced to find an intermediate correspondence representation.

	\section{METHOD}
	\label{sec:method}
	
	We first define our problem as a regression task~\cite{belagiannis2015robust}, followed by the objective function. Afterwards, we describe the proposed network architecture for point cloud registration.
	The problem definition is formulated based on \cite{engel2019landmark} and \cite{liu2019flownet3d}.
	
	\subsection{Problem Definition}
	\label{sec:problem_definition}
	
	Input to our network are two unsorted point sets $\mathcal{T} = \{a_i\}_{i=1}^{n_a}$ and $\mathcal{S} = \{b_i\}_{i=1}^{n_b}$ of arbitrary size $n_a$ and $n_b$ called template and source point cloud, respectively. Further, $a_i = \{\bm{x}_i, \bm{f}_i\}$ and $b_i = \{\bm{y}_i, \bm{g}_i\}$ are individual points, where $\bm{x}_i, \bm{y}_i \in \mathbb{R}^3$ denote $XYZ$ coordinates of points and $\bm{f}_i, \bm{g}_i \in \mathbb{R}^c$ denote optional input features like LiDAR intensities ($c=1$) or surface normals ($c=3$).
	The objective of registration is to find the linear transformation $\bm{T} \in \mathrm{SE}(3)$, consisting of translation $\bm{t} \in \mathbb{R}^3$ and rotation $\bm{R} \in \mathrm{SO}(3)$ in three-dimensional space that aligns the $XYZ$ coordinates of $\mathcal{T}$ and $\mathcal{S}$.
	The result of our training process is a mapping $h_{\bm{\theta}}(\mathcal{T}, \mathcal{S}) \rightarrow \bm{T}$ parameterized by $\bm{\theta}$, which is learned from training data.
	
	For LiDAR odometry, point clouds captured from two consecutive time steps are denoted $\mathcal{T}_t$ and $\mathcal{T}_{t + 1}$.
	The transformation $\bm{T}_t$ describes the motion between corresponding sensor poses $\bm{P}_t, \bm{P}_{t+1} \in \mathrm{SE}(3)$ with $\bm{P}_{t + 1} = \bm{P}_t \bm{T}_t$.
	The initial pose for odometry estimation is $\bm{P}_0 = \bm{I}$.

	\begin{figure*}
		\centerline{
			\subfigure[Network architecture]{
				\includegraphics[width=0.7\linewidth]{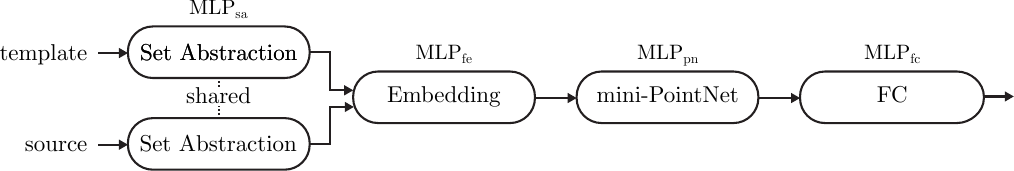}
				\label{fig:network_architecture}
			}
		}
		\centerline{
			\subfigure[Set Abstraction]{
				\includegraphics[height=3.3cm]{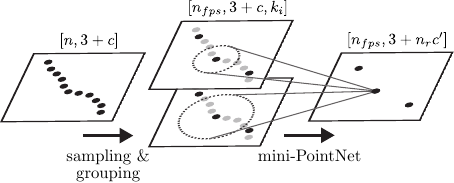}
				\label{fig:set_abstraction}
			}
			\hfil
			\subfigure[Flow Embedding]{
				\includegraphics[height=3.3cm]{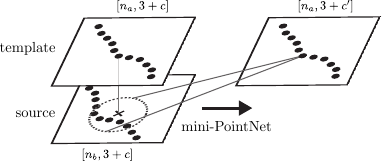}
				\label{fig:flow_embedding}
			}
		}
		\caption{Network architecture overview.
			The proposed architecture \subref{fig:network_architecture} first processes template and source point clouds with set abstraction, arranged as Siamese network with shared weights. The set abstraction \subref{fig:set_abstraction} subsamples the point cloud, performs multi-scale grouping and applies mini-PointNets on each group.
			The point clouds are combined using flow embedding \subref{fig:flow_embedding}, which groups all surrounding source points for each template point and applies mini-PointNets to obtain flow features.
			The resulting template points with flow features are then fed into a mini-PointNet with subsequent fully connected layers for predicting the alignment as dual-quaternion.
		}
		\label{fig:architecture}
	\end{figure*}

	\subsection{Objective}
	\label{sec:objective}
	
	The learning objective is to minimize the difference between predicted transformation $\bm{\hat T}$ and ground truth transformation $\bm{T}$.
	Usually, translations are described using a vector in $\mathbb{R}^3$, while for rotations multiple representations, e.g. Euler angles, $\mathrm{SO}(3)$ rotation matrices or quaternions, exist \cite{kendall2017geometric}.
	We chose to use dual-quaternions $\bm{\sigma} = \bm{p} + \epsilon \bm{q}$ described by Schneider et al. \cite{schneider2017regnet} for representing translation and rotation, consisting of the real part $\bm{p}$ and the dual part $\bm{q}$.
	The parts $\bm{p}$ and $\bm{q}$ are both quaternions defined as extended complex numbers $w + x\bm{i} + y\bm{j} + z\bm{k}$, where $w$, $x$, $y$ and $z$ are real numbers and $\bm{i}$, $\bm{j}$ and $\bm{k}$ are the imaginary components \cite{kenwright2012beginners}.
	Similar to the quaternion representation for rotation, the real part $\bm{p}$ contains the rotational information with values within $[-1, 1]$ and is usually normalized to $\left\|\bm{p}\right\| = 1$.
	The dual part $\bm{q}$ contains rotational and translational information without a specific range.
	More details on dual-quaternions can be found in \cite{kenwright2012beginners}.
	
	Based on the dual-quaternion representation, the dual loss is given by
	\begin{equation}
		L_{dual} = \mathbb{E} \left[ \left\|\bm{q} - \bm{\hat q} \right\|^2 \right]
	\end{equation}
	and the real loss by
	\begin{equation}
		L_{real} = \mathbb{E} \left[ \left\|\bm{p} - \frac{\bm{\hat p}}{\left\| \bm{\hat p}\right\|} \right\|^2 \right] \text{,}
	\end{equation}
	normalizing the predicted real part $\bm{\hat p}$ to a valid rotation quaternion \cite{kendall2017geometric, schneider2017regnet}.
	Since the dual part has no specific range, we compensate this imbalance by scaling the loss value with a factor $\beta$, which results in the combined loss function
	\begin{equation}
		L = \beta L_{real} + L_{dual} \text{.}
	\end{equation}
	We intentionally refrain from using homoscedastic uncertainty weighting as described in \cite{kendall2017geometric}, enabling us to manually set a higher weight on the rotation prediction for odometry estimation.

	\subsection{Network Architecture}
	\label{sec:network_architecture}
	
	\begin{figure}[b]
		\centerline{\includegraphics[width=0.65\linewidth]{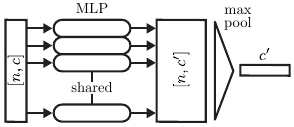}}
		\caption{Mini-PointNet architecture \cite{qi2017pointnet}. The module applies MLPs with shared weights on each input point and then performs max pooling over the point dimension in order to obtain a single feature vector.}
		\label{fig:point_net}
	\end{figure}
	
	In the following section, we describe our network architecture.
	An overview is given in Fig.~\ref{fig:network_architecture}.
	For subsampling and local feature extraction, the input point clouds are initially processed by the set abstraction \cite{qi2017pointnetpp}, which is arranged in a Siamese architecture with shared weights.
	Afterwards, the point clouds are merged with a flow embedding module \cite{liu2019flownet3d}.
	The flow embedding groups all surrounding source points for each template point in order to generate features describing the point flow between both point clouds.
	The resulting point cloud with added flow features is fed into a mini-PointNet with subsequent fully connected (FC) layers.
	In order to enforce valid predictions, the real part $\bm{p}$ of the dual-quaternion output is restricted using the sigmoid activation function for $w$ and the tanh activation function for $x$, $y$ and $z$ on the output layer.
	For further processing steps or evaluations, the predicted dual-quaternion is converted into the transformation matrix $\bm{T}$, consisting of translation $\bm{t}$ and rotation $\bm{R}$.
	
	The mini-PointNet structure \cite{qi2017pointnet} is used multiple times within our architecture.
	It is implemented as nonlinear function $h_p: \mathbb{R}^{c} \rightarrow \mathbb{R}^{c^{\prime}}$, realized as multi-layer perceptron (MLP), with subsequent element-wise max pooling:
	\begin{equation}
		\bm{f}^{\prime} = \MAX \{ h_p(\bm{f}) \} \text{.}
	\end{equation}
	The mini-PointNet is able to process an arbitrary number of points of dimension $c$ to a single feature vector of dimension $c^{\prime}$,
	as shown in Fig.~\ref{fig:point_net}.

	\textbf{Set Abstraction.}
	The set abstraction module is used to subsample the input points and add local features containing information about the surrounding point structure.
	We use a single set abstraction step from PointNet++ \cite{qi2017pointnetpp} with multi-scale grouping (MSG).
	The input point cloud is subsampled from $n$ to $n_{fps}$ points using iterative farthest point sampling (FPS).
	For each sample $a_j = \{\bm{x}_j, \bm{f}_j\}$, all surrounding points $a_k = \{\bm{x}_k, \bm{f}_k\}$ within radius $r_l \in \{r_i\}_{i=1}^{n_r}$ are grouped including the central point $a_j$.
	We define the distance vectors $\bm{d}_{k,j} = \bm{x}_k - \bm{x}_j$, which model the element-wise distance between central points $\bm{x}_k$ and surrounding points $\bm{x}_j$.
	The distance vectors $\bm{d}_{k,j}$ and features $\bm{f}_k$ are concatenated and fed into a mini-PointNet with nonlinear function $h_{sa}^{(l)}: \mathbb{R}^{3 + c} \rightarrow \mathbb{R}^{c^{\prime}}$, realized as $\text{MLP}_{\text{sa}}^{\text{(l)}}$, and element-wise max pooling.
	Following \cite{liu2019flownet3d}, this can be written as
	\begin{equation}
		\bm{f}_{j, l}^{\prime} = \MAX\limits_{\{k | \left\| \bm{d}_{k,j} \right\| < r_l\}} \{ h_{sa}^{(l)}(\bm{d}_{k,j}, \bm{f}_k) \} \text{.}
	\end{equation}
	Performing this for multiple radii $r_l$, the set abstraction result for each FPS sample is $a_j^{\prime} = \{\bm{x}_j, \bm{f}_j^{\prime}\}$, where $\bm{f}_j^{\prime} = \{ \bm{f}_{j, l}^{\prime} \}_{l = 1}^{n_r}$ denotes the concatenation for $n_r$ radii in multi-scale grouping.
	Thus, the input point cloud with dimensions $\left[n, 3 + c\right]$ is subsampled and transformed into a point cloud with dimensions $\left[n_{fps}, 3 + n_r c^{\prime}\right]$.
	See Fig.~\ref{fig:set_abstraction} for a visualization of the set abstraction.

	\textbf{Flow Embedding.}
	For merging the point clouds, we use the flow embedding module of \cite{liu2019flownet3d}.
	Flow embedding takes two point clouds $\mathcal{A} = \{ a_i \}_{i = 1}^{n_a}$ and $\mathcal{B} = \{b_i\}_{i = 1}^{n_b}$ with $a_i = \{\bm{x}_i, \bm{f}_i\}$ and $b_i = \{\bm{y}_i, \bm{g}_i\}$ as input.
	In our case, the point clouds are the result of the set abstraction of template point cloud $\mathcal{T}$ and source point cloud $\mathcal{S}$.
	Due to various causes like noise, missing measurements or occlusion, $\mathcal{B}$ does not always contain direct correspondences for each point in $\mathcal{A}$.
	Instead of searching for corresponding points, we can alternatively compare each point of $\mathcal{A}$ with all surrounding points in $\mathcal{B}$ for estimating the flow.
	It is important to note that this is only possible for small relative transformations.
	Thus, in case of large transformations a coarse pre-registration is required.
	
	The flow embedding is realized as follows:
	For each point $a_i$ of the first point cloud $\mathcal{A}$, all surrounding points $b_j$ of the second point cloud $\mathcal{B}$ within a certain radius $r$ are grouped.
	The distance vector $\bm{y}_j - \bm{x}_i$ and the features $\bm{f}_i$ and $\bm{g}_j$ are concatenated and fed into a mini-PointNet with nonlinear function $h_{fe}: \mathbb{R}^{3 + 2c} \rightarrow \mathbb{R}^{c^{\prime}}$, realized as $\text{MLP}_{\text{fe}}$, and element-wise max pooling.
	Similar to the set abstraction, this can be written as 
	\begin{equation}
		\bm{e}_i^{\prime} = \MAX\limits_{\{j | \left\| \bm{y}_j - \bm{x}_i \right\| < r\}} \{ h_{fe}(\bm{y}_j - \bm{x}_i, \bm{f}_i, \bm{g}_j) \} \text{,}
	\end{equation}
	resulting in $a_i^{\prime} = \{\bm{x}_i, \bm{e}_i^{\prime}\}$ as flow output for each point $a_i$.
	The concept of flow embedding is shown in Fig.~\ref{fig:flow_embedding}.
	Concatenating the features $\bm{f}_i$ and $\bm{g}_j$ instead of passing the feature distance vector $\bm{g}_j -\bm{f}_i$ was proven as a better approach in \cite{liu2019flownet3d}, since the network is able to learn an effective distance function for features.
	
	\textbf{Output.}
	The merged point cloud with flow features is reduced to a global feature vector using a mini-PointNet with $\text{MLP}_{\text{pn}}$.
	This removes the necessity for extracting explicit correspondences between both point clouds, since the flow features can carry more information about the point flow than explicit correspondences or correspondence weights.
	The global feature vector is finally fed into a fully connected $\text{MLP}_{\text{fc}}$ with 8 outputs for predicting a dual-quaternion.

	\section{IMPLEMENTATION DETAILS}
	\label{sec:implementation_details}
	
	In contrast to the original PointNet++, we only use a single set abstraction layer with $n_{fps}$ FPS samples and two MSG radii in order to maintain a short inference time.
	For the radius search in set abstraction (SA) and flow embedding (FE), we have to provide a maximum number of samples $n_{s,sa}$ and $n_{s,fe}$ in addition to the radii.
	The parameters $n_{s,sa}$ and $n_{s,fe}$ were chosen based on the distribution of our training data.
	The radii for flow embedding were selected based on the maximum possible offset of the points for given ground truth transformations.
	An overview of the network parameters is given in Table~\ref{tab:architecture_specs}.
	
	\begin{table}
		\renewcommand{\arraystretch}{1.3}
		\centering
		\caption{Architecture hyperparameters for KITTI and ModelNet40.}
		\begin{tabular}{ccc}
			\hline
			Parameter & KITTI & ModelNet40 \\
			\hline
			$n_{fps}$ & 1024 & 512 \\
			SA radii & 0.5, 1.0 & 0.05, 0.1 \\
			$n_{s,sa}$ & 512, 1024 & 256, 512 \\
			FE radius & 10.0 & 0.2 \\
			$n_{s,fe}$ & 15 & 30 \\
			Loss $\beta$ & 200 & 1 \\
			$\text{MLP}_{\text{sa}}$ & \multicolumn{2}{c}{[16, 16, 32]} \\
			$\text{MLP}_{\text{fe}}$ & \multicolumn{2}{c}{[128, 128, 256]} \\
			$\text{MLP}_{\text{pn}}$ & \multicolumn{2}{c}{[256, 512, 512, 1024]} \\
			$\text{MLP}_{\text{fc}}$ & \multicolumn{2}{c}{[512, 256, 8]} \\
			\hline
		\end{tabular}
		\label{tab:architecture_specs}
	\end{table}
	
	We use weight decay in combination with data augmentation as regularization.
	For augmentation, we can simply apply random translations and rotations to the source point clouds and ground truth transformations.
	
	For LiDAR odometry of sequential measurements, we can reduce the inference time by reusing the set abstraction output of the previous registration pair.
	When predicting the transformation from time $t_i$ to $t_{i+1}$, we already have the set abstraction of point cloud $t_i$ available from the previous inference of transformation $t_{i-1}$ to $t_i$.
	On the first sample $t_0$, we only perform the set abstraction, since we have no second point cloud available.
	
	\section{EXPERIMENTS}
	\label{sec:experiments}
	
	We conduct two experiments with point clouds from different sources for demonstrating the versatility of our architecture.
	The KITTI odometry dataset \cite{geiger2012autonomous} is used to show the ability of our architecture to register noisy real world data with a long range and high numbers of points, including dynamic objects, which adversely affect the registration.
	Besides that, we use an adapted version of the ModelNet40 dataset \cite{wu20153d} to evaluate the performance on ideal ground truth data with artificial transformations and noise.
	
	For evaluation, the translation error $t_{err}$ and rotation error $r_{err}$ between ground truth transformation $(\bm{R}_{gt}, \bm{t}_{gt})$ and prediction $(\bm{R}_{pred}, \bm{t}_{pred})$ are calculated as Euclidean distance in \si{\meter} and chordal distance \cite{hartley2013rotation} in \si{deg}, respectively:
	\begin{align}
		t_{err} &= \left\|\bm{t}_{pred} - \bm{t}_{gt} \right\|_2 \label{eq:t_err} \text{,}\\
		r_{err} &= 2 \arcsin \left( \frac{\left\| \bm{R}_{pred} - \bm{R}_{gt} \right\|_F}{\sqrt{8}} \right) \text{.} \label{eq:r_err}
	\end{align}
	Alternatively, the methods can be evaluated using the root mean square errors (RMSEs) $t_{rmse}$ and $r_{rmse}$ in \si{\metre} and \si{\deg} of translation vector and rotation in Euler angles.
	
	We use ICP based on the Open3D library \cite{zhou2018open3d} and \mbox{G-ICP} based on the original reference implementation \cite{segal2010generalized} for comparison.
	The maximum distance between corresponding points for ICP is set according to the radius used for our flow embedding.
	All methods are tested on a machine with a 3.70\,GHz CPU, 8\,GB RAM and a GPU with 11\,GB memory.
	Our approach is implemented in PyTorch.

	\subsection{KITTI Odometry}
	
	The KITTI odometry dataset \cite{geiger2012autonomous} consists of 22 sequences with stereo gray-scale and color camera images and point clouds captured by a Velodyne laserscanner.
	The data is recorded in different road environments including static and dynamic objects.
	The LiDAR point clouds are scanned with \SI{10}{\Hz}.
	Ground truth trajectories for sequences 00-10 with 23\,201 scans are publicly available for training and evaluation, trajectories for sequences 11-21 with 20\,351 scans are hidden for benchmark purposes.
	Each LiDAR measurement consists of 64 scan layers with approximately 100\,000 points.
	To reach an inference time of less than \SI{100}{\milli\second}, we reduce the scan size by removing every other point.
	
	It is important to note that the provided ground truth trajectories are partially inaccurate, which has also been observed by other authors \cite{lu2019deepvcp}.
	For compensating incorrect labels, we implement an additional augmentation step on top of random transformations and noise.
	We create additional scan pairs by duplicating the template point clouds and applying random transformations and noise for creating the source point clouds.
	Although this creates similar scan pairs, the associated transformation labels reliably involve no error.
	
	\begin{table}
		\begin{center}
			\begin{threeparttable}
				\renewcommand{\arraystretch}{1.3}
				\setlength{\tabcolsep}{1.1em}
				\caption{KITTI odometry results for sequences 04 and 10.}
				\label{tab:kitti_results_odometry}
				\begin{tabular}{lcccc}
					\hline
					& \multicolumn{2}{c}{04} & \multicolumn{2}{c}{10} \\
					Method & $t_{rmse}$ & $r_{rmse}$ & $t_{rmse}$ & $r_{rmse}$ \\
					\hline
					DeepPCO\tnote{*} ~\cite{wang2019deeppco} & \textbf{0.0263} & 0.0305 & 0.0247 & 0.0659 \\
					Ours & 0.0264 & \textbf{0.0283}& \textbf{0.0202} & \textbf{0.0475} \\
					\hline
				\end{tabular}
				\begin{tablenotes}[flushleft]
					\item Both methods were trained with sequences 00-03 and 05-09.
					\item Units for $r_{rmse}$ and $t_{rmse}$ are \si{deg} and \si{\metre}, respectively.
					\item[*] Values taken from \cite{wang2019deeppco}.
				\end{tablenotes}
			\end{threeparttable}
		\end{center}
	\end{table}
	
	\textbf{LiDAR Odometry.} 
	For training LiDAR odometry, translation and rotation augmentation noise are Gaussian distributed with standard deviation [\SI{0.2}{\meter}, \SI{0.02}{\meter}, \SI{0.02}{\meter}] and [\SI{0.1}{\degree}, \SI{0.1}{\degree}, \SI{1.0}{\degree}] in Euler angles.
	The point noise for training is Gaussian distributed with standard deviation \SI{0.01}{\meter}.
	Various splits of the ground truth sequences 00-10 into train and test set and different metrics are used in literature.
	
	Li et al. train LO-Net \cite{li2019lonet} with 00-06 and test with 07-10 using the KITTI odometry evaluation metrics \cite{geiger2012autonomous}, which consist of the average relative translation and rotation errors in \si{\percent} and \si{\degree\per100\meter} for segments of \SI{100}{\meter} to \SI{800}{\meter}.
	We compare our architecture to the performance metrics of LO-Net without mapping and to LOAM \cite{zhang2014loam}, as given in \cite{li2019lonet}.
	Our architecture yields an average error of \SI{2.18}{\percent} and \SI{0.93}{\degree\per100\meter}.
	LO-Net with \SI{1.75}{\percent} and \SI{0.79}{\degree\per100\meter}, and LOAM with \SI{1.15}{\percent} and \SI{0.50}{\degree\per100\meter} achieve a better performance.
	This is not surprising, since our learning objective is only focused on registration of consecutive scans instead of odometry for longer sequences.
	Furthermore, both LO-Net and LOAM additionally exploit the sorted and grid-like structure of LiDAR measurements, which is intentionally not used by our method.
	
	In contrast, DeepPCO \cite{wang2019deeppco} is trained with 00-03, 05-09 and tested with 04, 10.
	They use the average RMSEs $t_{rmse}$ and $r_{rmse}$ over all scan pairs as metrics.
	The results are shown in Table~\ref{tab:kitti_results_odometry}.
	For all metrics except $t_{rmse}$ of sequence 04, where the results are almost equal, we achieve a better performance.
	Our average and maximum inference times are \SI{55}{\milli\second} and \SI{76}{\milli\second}, which is far below the upper limit of \SI{100}{\milli\second} for online processing.
	
	\begin{table}
		\begin{center}
			\begin{threeparttable}
				\renewcommand{\arraystretch}{1.3}
				\setlength{\tabcolsep}{0.4em}
				\caption{KITTI results with artificial transformations.}
				\label{tab:kitti_results_artificial}
				\begin{tabular}{lccccc}
					\hline
					& \multicolumn{2}{c}{Rot. Error (\si{deg})} & \multicolumn{2}{c}{Trans. Error (\si{\metre})} & Time (\si{\second}) \\
					Method & Mean & Max & Mean & Max & Mean \\
					\hline
					ICP Point2Point \cite{besl1992method} & 0.088 & 0.691 & 0.177 &  4.703 & 0.61 \\
					ICP Point2Plane \cite{besl1992method} & 0.079 & 0.538 & 0.140 &  4.784 & 0.98 \\
					G-ICP \cite{segal2010generalized} & \textbf{0.029} & \textbf{0.215} & 0.109 & 4.746 & 3.21 \\
					3DFeat-Net\tnote{*} ~\cite{yew20183dfeatnet} & 0.199 & 2.428 & 0.116 & 4.972 & 15.02 \\
					DeepVCP\tnote{*} ~\cite{lu2019deepvcp} & 0.164 & 1.212 & \textbf{0.071} & \textbf{0.482} & 2.3 \\
					Ours & 0.053 & 0.389 & 0.080 & 1.239 & \textbf{0.08} \\
					\hline
				\end{tabular}
				\begin{tablenotes}[flushleft]
					\item Sequences 00-07 were used for training, sequences 08-10 for testing.
					\item Error metrics $r_{err}$ \eqref{eq:r_err} and $t_{err}$ \eqref{eq:t_err} are used.
					\item[*] Values taken from \cite{lu2019deepvcp}.
				\end{tablenotes}
			\end{threeparttable}
		\end{center}
	\end{table}
	
	\textbf{Artificial Transformations.}
	In contrast to the previous evaluation for odometry, DeepVCP \cite{lu2019deepvcp} is focused on the general registration of LiDAR measurements, independent of the odometry task.
	Thus, they sample the sequences at 30 frame intervals for templates and use all subsequent scans within \SI{5}{\meter} as sources.
	The artificial samples are created by removing the ground truth transformation calculated from the trajectory and applying unbiased, uniformly distributed transformations with maximum translation \SI{1.0}{\meter} and rotation \SI{1.0}{\degree}.
	As metrics, mean and maximum angular error $r_{err}$ as well as translation error $t_{err}$ are used.
	The dataset is split into sequences 00-07 for training and 08-10 for testing.
	The results are shown in Table~\ref{tab:kitti_results_artificial}.
	
	Our proposed architecture has by far the shortest average run-time of all compared methods.
	The average inference time of \SI{81}{\milli\second} is higher compared to the odometry task, since we have to perform the set abstraction on both clouds for non-consecutive measurements.
	For rotation, G-ICP achieves the best results, while for translation, DeepVCP has the lowest average and maximum errors.
	In both cases, our proposed architecture is the second best approach with only a small gap to the best method.
	Considering this and the comparatively high run-time of the other methods, our architecture achieves state-of-the-art results.
	The high maximum translation error in all methods is most likely a result of the inaccurate ground truth trajectories mentioned previously.
	
	\begin{table}
		\begin{center}
			\begin{threeparttable}
				\renewcommand{\arraystretch}{1.3}
				\setlength{\tabcolsep}{0.3em}
				\caption{ModelNet40 results on test data from unseen categories.}
				\label{tab:modelnet40_results_unseen}
				\begin{tabular}{lccccc}
					\hline
					& \multicolumn{2}{c}{Rot. Error (\si{deg})} & \multicolumn{2}{c}{Trans. Error (\si{unit})} & Time (\si{\milli\second}) \\
					Method & Mean & Std. Dev. & Mean & Std. Dev. & Mean \\
					\hline
					ICP Point2Point \cite{besl1992method} & 0.472 & 0.270 & 0.0092 & 0.0051 & \phantom{0}86.3 \\
					ICP Point2Plane \cite{besl1992method} & 0.559 & 0.508 & 0.0076 & 0.0038 & \phantom{0}83.8 \\
					G-ICP \cite{segal2010generalized} & 0.695 & 0.374 &  0.0154 & 0.0071 & 101.9 \\
					Ours & \textbf{0.328} & \textbf{0.192} & \textbf{0.0055} & \textbf{0.0028} & \textbf{\phantom{0}14.6} \\
					\hline
				\end{tabular}
				\begin{tablenotes}[flushleft]
					\item Results for Gaussian distributed noise with standard deviation \SI{0.04 }{units}. Error metrics $r_{err}$ \eqref{eq:r_err} and $t_{err}$ \eqref{eq:t_err} are used.
				\end{tablenotes}
			\end{threeparttable}
		\end{center}
	\end{table}
	
	\subsection{ModelNet40}
	
	\begin{figure}[t]
		\centering
		\includegraphics[width=0.95\linewidth]{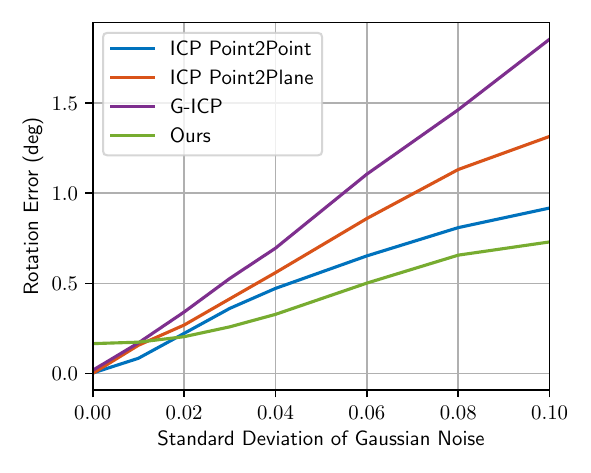}
		\caption{ModelNet40 evaluation results on test data from unseen categories and increasing standard deviation of the Gaussian noise. Error metric $r_{err}$ \eqref{eq:r_err} is used.}
		\label{fig:modelnet40_unseen_rerr}
	\end{figure}
	
	\begin{figure}[b]
		\centering
		\includegraphics[width=0.85\linewidth]{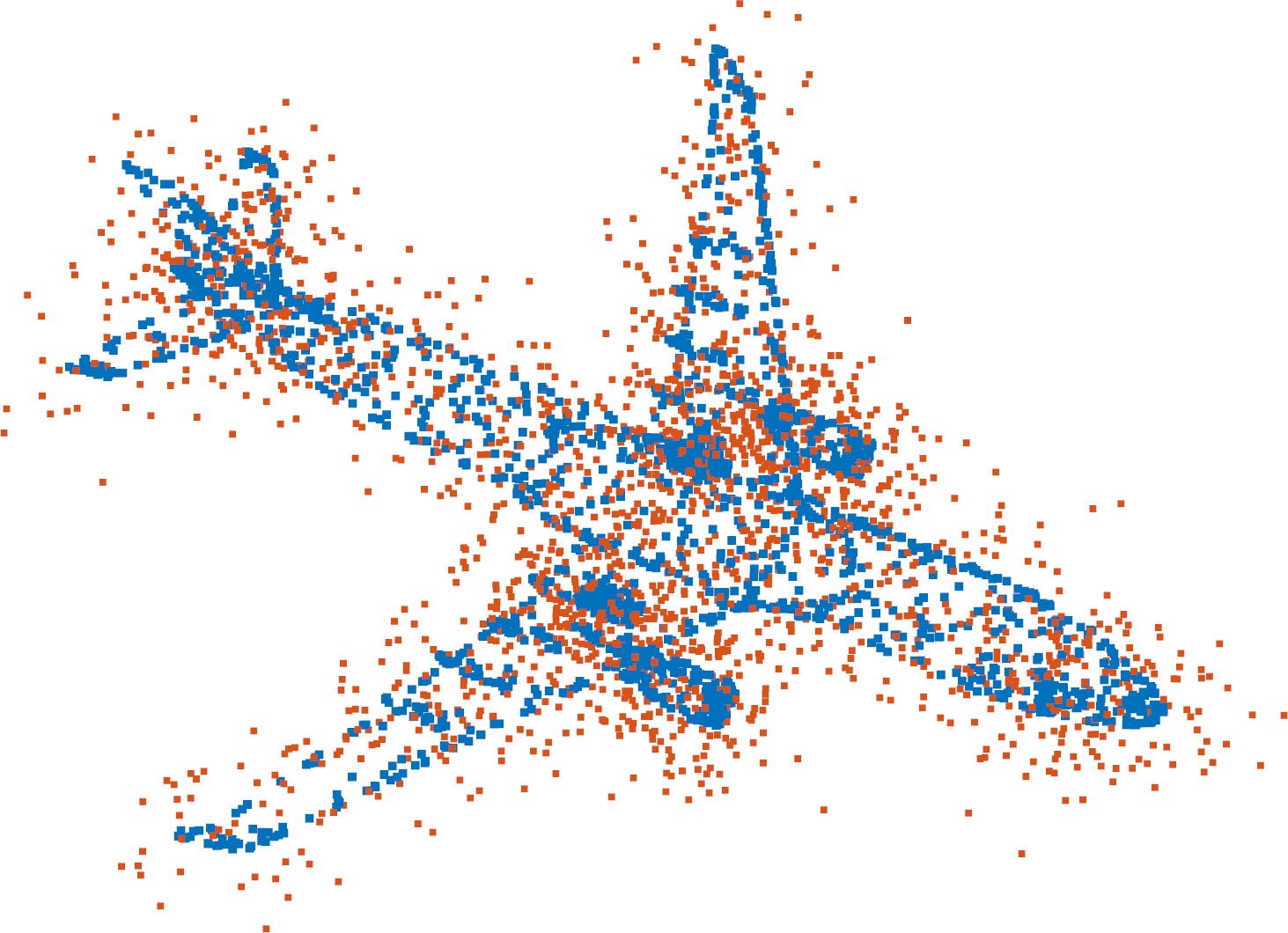}
		\caption{ModelNet40 \cite{wu20153d} example of the airplane category.
			The subsampled point cloud with 2048 points is shown in blue. 
			The point cloud in red has additional Gaussian distributed noise with standard deviation \SI{0.04}{units}.}
		\label{fig:modelnet40_plane}
	\end{figure}
	
	The ModelNet40 \cite{wu20153d} dataset provides over 12\,000 CAD models of 40 different shape categories like airplane, chair, person or table.
	Since the CAD vertices are non-uniformly distributed, Qi et al. \cite{qi2017pointnetpp} provide a version with 10\,000 uniformly sampled surface points and normals per model.
	By applying random transformations and point noise to the point clouds, this dataset can be used for the evaluation of point cloud registration.
	We use FPS to subsample the data to 2048 points in order to achieve a manageable run-time for the ICP variants.
	An example is shown in Fig.~\ref{fig:modelnet40_plane}.
	
	Since we only consider fine registration, we apply random transformations with translation $[0, 0.1]$\,\si{units} and rotation angles $[0, 5]$\,\si{deg} about arbitrarily chosen axes during training and testing, which is significantly smaller than in \cite{aoki2019pointnetlk}.
	Furthermore, Gaussian distributed random noise with standard deviation \SI{0.02}{units} is added to template and source points during training.
	In contrast to \cite{aoki2019pointnetlk}, we add noise to template and source points, since our intended applications include the registration of real world measurements, where both point clouds are noisy.
	For testing, we perform multiple runs using Gaussian distributed random noise with standard deviations from \SI{0.0}{units} to \SI{0.1}{units}.
	
	Our architecture is trained and tested with point clouds from 20 different shape categories.
	To examine our ability to generalize and for fair comparison with classic algorithms, we further perform tests on 20 previously unseen shape categories.
	All methods are tested on the same point clouds, noise data and transformations.
	
	Since we did not notice a performance difference between seen and unseen shape categories, our network proves to generalize well instead of learning shape-specific features.
	Table~\ref{tab:modelnet40_results_unseen} shows results for unseen categories and Gaussian distributed noise with standard deviation \SI{0.04}{units}.
	The rotation error over increasing noise standard deviations is given in Fig.~\ref{fig:modelnet40_unseen_rerr}.
	For low noise levels below \SI{0.01}{units}, all ICP variants achieve better results, since template and source point clouds are more similar.
	Starting at standard deviation \SI{0.02}{units}, our architecture clearly outperforms the compared methods considering error average and standard deviation, and run-time.
	Although we only used a noise level of \SI{0.02}{units} during training, we still achieve good performance for higher noise levels.
	Again, this demonstrates the powerful generalization capability of our architecture.
	
	\section{CONCLUSION}
	\label{sec:conclusion}
	
	In this work, we have presented DeepCLR, a novel architecture for point cloud registration based on PointNet and PointNet++.
	In contrast to previous approaches, we provide an end-to-end architecture based on radius search and mini-PointNets for merging the input point clouds without extracting explicit point correspondences.
	We have evaluated our method on LiDAR measurements from the KITTI odometry dataset and on model data created from ModelNet40.
	On both datasets, our architecture achieves state-of-the-art results while providing the lowest run-time of the compared methods.
	Furthermore, on noisy ModelNet40 data our architecture clearly outperforms classic algorithms like ICP.
	In future work, we would like to explore pretraining of the set abstraction for improved local features.

	
	


	

	\section*{ACKNOWLEDGMENT}
	
	This research was accomplished within the project UNICARagil (FKZ 16EMO0290).
	We acknowledge the financial support for the project by the Federal Ministry of Education and Research of Germany (BMBF).
	
	
	
	\printbibliography
	
\end{document}